\documentclass[letterpaper, 10 pt, conference]{ieeeconf}

\IEEEoverridecommandlockouts 
\usepackage{amsmath} 
\usepackage{amssymb}  
\usepackage{tabularx, graphicx,
  amsmath,amsfonts,siunitx,pifont,amssymb,subcaption}
\usepackage{xcolor, bm, float, multirow, multicol, courier, mathtools, url}
\usepackage{footmisc}
\usepackage[linesnumbered,ruled,vlined]{algorithm2e} \pdfminorversion=4
\let\labelindent\relax \usepackage{enumitem} \usepackage{graphicx}

\usepackage{siunitx}
\DeclareSIUnit[product-units = single]{\pixel}{px}

\usepackage{hyperref}
\usepackage{siunitx}

\usepackage{pdfpages}
\usepackage{cite}
\usepackage{algpseudocode}
\usepackage{float}
\usepackage{colortbl}
\usepackage{dblfloatfix}    
\usepackage{balance}

\def\secref#1{Sec.~\ref{#1}}
\def\figref#1{Fig.~\ref{#1}}
\def\tabref#1{Tab.~\ref{#1}}
\def\eqref#1{Eq.~(\ref{#1})}

\title{\LARGE \bf Multi-Camera LiDAR Inertial Extension to the Newer College
Dataset}

\author{Lintong Zhang, Marco Camurri, David Wisth and Maurice Fallon
	\thanks{The authors are with the Oxford Robotics Institute, University of Oxford, UK.
		{\tt\small \{lintong, mcamurri, davidw, mfallon\}@robots.ox.ac.uk}}%
}

\begin{document}

\setlength{\abovedisplayskip}{4pt}
\setlength{\belowdisplayskip}{4pt}

\maketitle \thispagestyle{empty} \pagestyle{empty}

\begin{abstract}

We present a multi-camera LiDAR inertial
dataset of \SI{4.5}{\kilo\meter} walking distance as an expansion of the Newer
College Dataset. The global shutter multi-camera device is hardware
synchronized with both the IMU and LiDAR, which is more accurate than the
original dataset with software synchronization. This dataset also provides
six Degrees
of Freedom (DoF) ground truth poses at LiDAR frequency (\SI{10}{\hertz}). Three
data
collections are described and an example use case of multi-camera
visual-inertial
odometry is demonstrated. This expansion dataset contains small and narrow
passages, large scale open spaces, as well as vegetated areas, to test
localization and mapping systems. Furthermore, some sequences present
challenging situations such as abrupt lighting change, textureless surfaces,
and aggressive motion. The dataset is available at:
\url{https://ori-drs.github.io/newer-college-dataset/}

\end{abstract}

\section{Introduction}

There has been rapid progress in the field of robotic autonomous navigation.
High-quality public datasets can propel research and development, allowing
consistent evaluation across different algorithms. Our recent Newer College
Dataset \cite{ramezani2020newer} features a stereo-inertial camera and a dense
LiDAR setup, and introduced a novel method of generating accurate high-frequency
ground truth poses. In line with the development of cutting-edge sensors, we
now present a new handheld device incorporating a more accurately synchronized
multi-camera device for challenging visual scenarios, and a wide field of view
128 channel LiDAR that can provide even denser point clouds per single scan.

Recently, many commercial robotic applications rely on state estimation systems
containing multi-camera configurations. Despite the fact, the vast majority of
datasets SLAM practitioners uses to benchmark their systems are based on either
monocular or stereo camera configurations. There
exists a gap in considering different types of sensing configurations among
industrial applications and academic research, and this dataset would hope to
serve such a purpose.

There are several monocular and stereo camera datasets for visual odometry and
SLAM purpose. Some of these datasets
contain IMU or LiDAR measurements. A notable example is the
KITTI dataset \cite{Geiger2013IJRR} that provides LiDAR, IMU, and stereo camera
data with GPS/INS ground truth.
EuRoC MAV\cite{Burri25012016} and TUM VI \cite{schuberttum} both offer 6 Degree
of Freedom (DoF)
ground truth poses, with data for IMU and stereo cameras, but not any LiDAR
sensors. For a more detailed comparison, we refer the reader to Table I in our
original paper \cite{ramezani2020newer}. However, there is very few public
dataset that offers multi-camera vision for odometry or SLAM related usage.

Within the field of computer vision, there are a number of public datasets
involving multiple cameras. EPFL-RLC \cite{EPFL_RLC} dataset used three static
HD cameras inside a building to track objects such as pedestrians. WoodScape
\cite{yogamani2019woodscape} was the first fisheye autonomous driving dataset
with 4 large Fields of View (FoV) cameras. It provided several categories of
information including segmentation, depth estimation, and bounding box. These
datasets can be used for a variety of computer vision-related research such as
object tracking, prediction, and classification, but not for autonomous
navigation tasks such as localisation and mapping.

To the best of our knowledge, the only dataset with similar sensors to ours is
the Hilti SLAM Challenge dataset \cite{helmberger2021hilti}. This dataset
focused on the construction site environment, providing sparse ground truth
poses, with just a few ground truth measurements per sequence. Most sequences
are collected in a ``stop and go'' fashion, where the total station produces a
measurement using a reflecting prism during the ``stop'' periods.

We would like to provide the research community with a comprehensive
multi-camera dataset that encompasses vision, LiDAR, inertial measurements,
along with high-frequency ground truth poses and accurate prior maps.

\begin{figure}
	\centering
	\includegraphics[width=\linewidth]{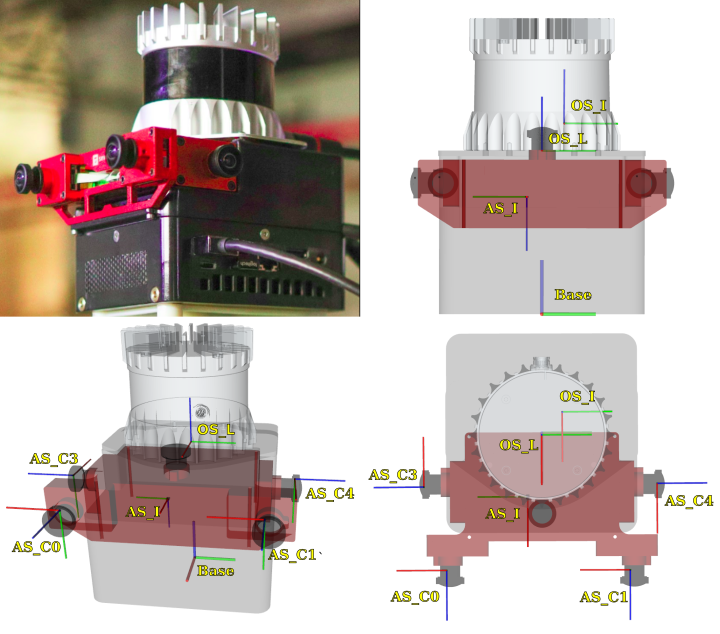}
	\caption{Our custom built handheld device is on the top left. The other
	three images are the URDF model with reference frames shown in front, isometric,
	and top-down views.}
	\label{fig:halo}
\end{figure}

\section{The Handheld Device}
\label{sec:platform}
The handheld multi-camera LiDAR inertial device is shown in \figref{fig:halo}.
The sensors are rigidly attached to a precisely 3D-printed base. A complete
URDF model of the device is available as an open source ROS
package\footnote{\url{https://github.com/ori-drs/halo\_description}} (see
\figref{fig:halo}). The Ouster LiDAR is directly mounted above the cameras for
a balanced and compact design, so the top facing camera was removed.
\tabref{table:HaloOverview} provides an overview of the various
sensors.

The multi-camera sensor in our device is the Alphasense Core Development Kit
from Sevensense Robotics AG. An FPGA within the Alphasense synchronizes the IMU
and four grayscale fisheye cameras – a
frontal stereo pair with an 11 cm baseline and two lateral cameras. Each camera
has a FoV of \SI{126 x 92.4}{\degree} and a resolution of
\SI{720 x 540}{\pixel}. This configuration also produces an overlapping FoV
between the front and side cameras of about \SI{36}{\degree}. The cameras and
the embedded cellphone-grade IMU operates at \SI{30}{\hertz} and
\SI{200}{\hertz}, respectively.
The Ouster OS-0 LiDAR has 128 beams and a 90 degree elevation FoV, which
provides a much denser point cloud than the OS1-64 used in the original dataset.
Both sensors are cutting-edge devices in mobile robotics
research and have recently been used in the DARPA Subterranean Challenge.

\figref{fig:halo} shows the various sensor frames with the following
abbreviations:
\begin{itemize}
	\item Base: The bottom centre of the printed computer case.
	\item OS{\_}I: The IMU coordinate system in the LiDAR.
	\item OS{\_}L: The LiDAR coordinate system where the point clouds are read.
	\item AS{\_}C0, 1, 3, 4: Camera optical frames.
	\item AS{\_}I: The IMU coordinate system in the Alphasense.
\end{itemize}

\begin{table}
	\centering
	\resizebox{\linewidth}{!}{%
		\begin{tabular}{llll}
			\hline
			\textbf{Sensor}&\textbf{Type}&
			\textbf{Rate}&\textbf{Characteristics}\\
			\hline
			\hline
			LiDAR & Ouster, OS0-128 & 10 Hz & 128 Channels, 50 m Range\\
			& & & 90$^{\circ}$ Vertical FoV\\
			& & &1024 Horizontal Resolution\\
			Cameras&Alphasense &30 Hz &Global shutter (Infrared)\\
			& & &720$\times$540\\
			LiDAR IMU&ICM-20948 &100 Hz &3-axis Gyroscope\\
			& & &3-axis Accelerometer\\
			Camera IMU&Bosch BMI085 &200 Hz &Sychronized with cameras\\
			\hline
		\end{tabular}
	}
	\caption{\small{Overview of the sensors in our handheld device.}}
	\label{table:HaloOverview}
\end{table}

\begin{figure}
	\centering
	\includegraphics[width=\linewidth]{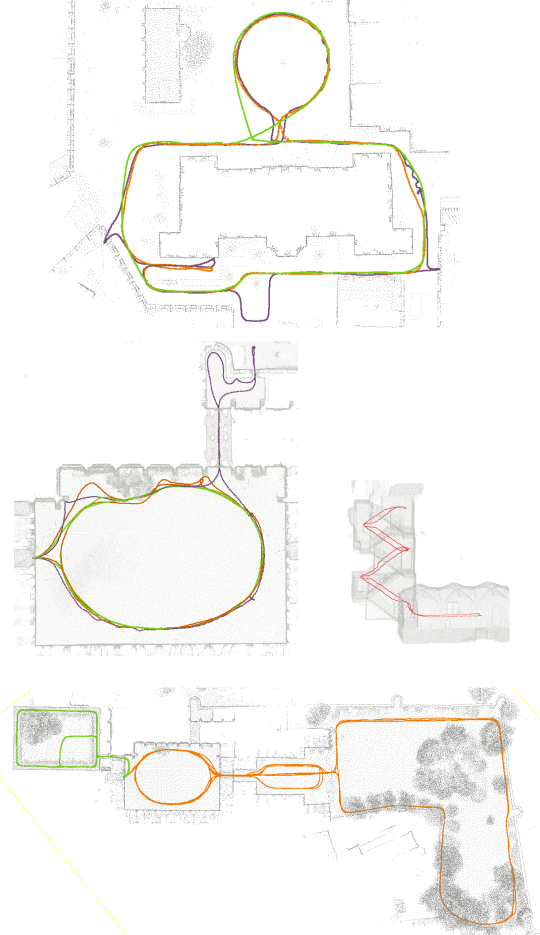}
	\caption{Trajectories for all collections. \textbf{Top:}
		Maths-Easy (green), Maths-Medium (orange), Maths-Hard (purple);
		\textbf{Mid left:} Quad-Easy (green), Quad-Medium (orange), Quad-Hard
		(purple); \textbf{Mid right:} Stairs; \textbf{Bot:} Cloister-Easy
		(green), Park (orange).}
	\vspace{-4mm}
	\label{fig:all_traj}
\end{figure}

\section{Data Collection}
\label{sec:dataCollection}
We have collected a variety of datasets at different speeds of walking and
turning. The three collections of datasets were gathered at different times of
the year and they contain some challenging aspects such as fast motions,
aggressive shaking, rapid lighting change, and textureless surfaces. The high
frequency ground truth trajectories are also provided using the same method in
the original paper, summarized in \secref{sec:groundTruth}.

The datasets contain different levels of difficulty and are organized
according to the aggressiveness of the motion and the type of scenes observed
by the cameras.

\textbf{Collection 1 - New College:}
\begin{itemize}
	\item \textit{Quad-Easy} (\SI{198}{\second}): Two loops in the quad area
	with a typical walking speed (\SI{247}{\meter}).
	\item \textit{Quad-Medium} (\SI{190}{\second}): Two loops of brisk walking
	with cameras pointing in different directions on different occasions
	(\SI{260}{\meter}).
	\item \textit{Quad-Hard} (\SI{187}{\second}): Fast walking with
	aggressive motion, approaches to the walls, and lighting changes
	(\SI{234}{\meter}).
	\item \textit{Stairs} (\SI{118}{\second}): Climbing up and down in a narrow
	stairway with the cameras subject to doors opening and textureless corridor
	walls (\SI{57}{\meter}).
\end{itemize}

\textbf{Collection 2 - New College:}
\begin{itemize}
	\item \textit{Cloister} (\SI{278}{\second}): Two loops of the cloister
	corridor and the cloister centre quad (\SI{429}{\meter}).
	\item \textit{Park} (\SI{1567}{\second}): Long experiment of the entire
	park and two quads with multiple loops. Route corresponds to the original dataset (\SI{2396}{\meter}).
\end{itemize}

\textbf{Collection 3 - Maths Institute:}
\begin{itemize}
	\item \textit{Maths-Easy} (\SI{216}{\second}): Outdoor large scale
	environment with typical walking speed (\SI{264}{\meter}).
	\item \textit{Maths-Medium} (\SI{176}{\second}): Brisk walking with cameras
	occasionally turned to face different directions (\SI{304}{\meter}).
	\item \textit{Maths-Hard} (\SI{243}{\second}): Fast walking with
	aggressive motions, textureless surfaces, rapid rotations and shaking of the
	device up to \SI{5.5}{\radian/\second}
	(\SI{321}{\meter}).
\end{itemize}

In \figref{fig:all_traj}, trajectories for each collection are overlaid on
the ground truth map using the method explained in the next section. A video
example displaying four camera steams and the point cloud of the cloister
sequence can be viewed at \url{https://youtu.be/tGXNSlmQOb0}.

\section{Ground Truth}
\label{sec:groundTruth}
\begin{figure}
	\includegraphics[width=\linewidth]{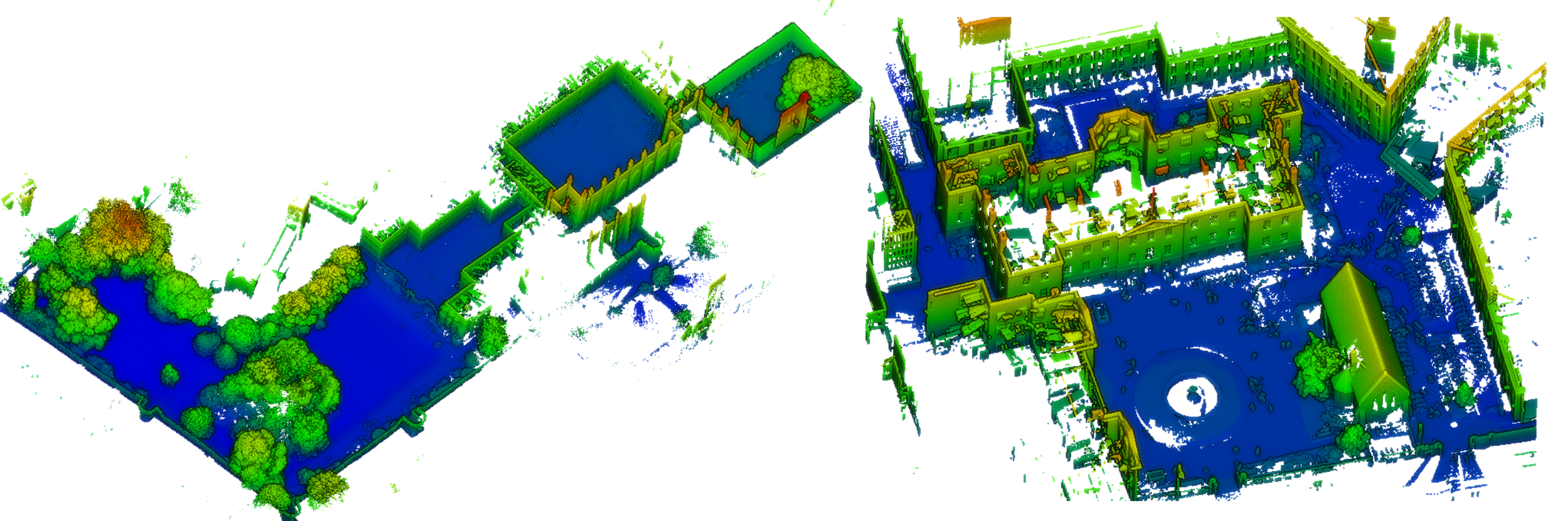}
	\caption{Ground truth and prior maps. \textbf{Left:} New College, Oxford
	(extended from the original Newer College dataset map);
	\textbf{Right:} Maths Institute, University of Oxford.}
	\label{fig:prior-map}
\end{figure}

The prior map and ground truth pose generation process use the same method
described in \cite{ramezani2020newer}. We use a survey-grade 3D imaging laser
scanner, Leica BLK360, to scan the entire environment. As shown in
\figref{fig:prior-map}, we further extend the prior map of New College and
provide an additional map for the Maths Institute. Ground truth poses are
determined by registering each undistorted LiDAR scan to the prior map using an
approach
based on the point-to-point Iterative Closest Point method. By processing each
scan at a lower speed, we can accurately correct the motion distortion using
integrated IMU measurement. The poses are expressed in the ``Base'' frame shown
in Sec. \ref{sec:platform}.

\section{Calibration}

Similarly to the original stereo camera LiDAR dataset, we use the open source
camera and IMU calibration toolbox Kalibr \cite{Rehder2016kalibr} to compute the
intrinsic calibration of the Alphasense cameras as well as their extrinsics. We
perform spatio-temporal calibration between the cameras and the two IMUs
embedded in the Alphasense and the Ouster sensor. The calibration target is
$6\times6$ April grids and each tag size is \SI{8.8}{\centi\meter}. The
calibration settings use the pinhole projection model with equidistant
distortion. All cameras are calibrated with the IMU, starting with the frontal
stereo cameras, then the individual lateral cameras. Since the collections were
carried out at different
times of the year, we provide a set of calibration files corresponding to each
dataset collection.

\subsection{Synchronization}
The recording computer, Intel NUC, is modified to support dual Ethernet ports
with hardware timestamp capability. The Ouster LiDAR and Alphasense are
synchronized with the NUC using the Precision Time Protocol (PTP), which
achieves sub-microsecond accuracy.

Further details about the Alphasense synchronization can be found at its Github
page\footnote{\url{https://github.com/sevensense-robotics/alphasense_core_manual}}.
The detailed synchronization procedure for Ouster LiDAR can be found in the
software
manual\footnote{\url{https://data.ouster.io/downloads/software-user-manual/software-user-manual-v2p0.pdf}}
 (Section 16 “PTP Quickstart Guide”).

\section{Example Usage}
To demonstrate one usage of the dataset and the advantage of multi-sensor
synchronization, we compared three visual-inertial odometry methods,
with reference to LiDAR-generated ground truth.

\begin{figure}
	\centering
	\includegraphics[width=\columnwidth, trim={0.25cm 0 0.25cm 0},
	clip]{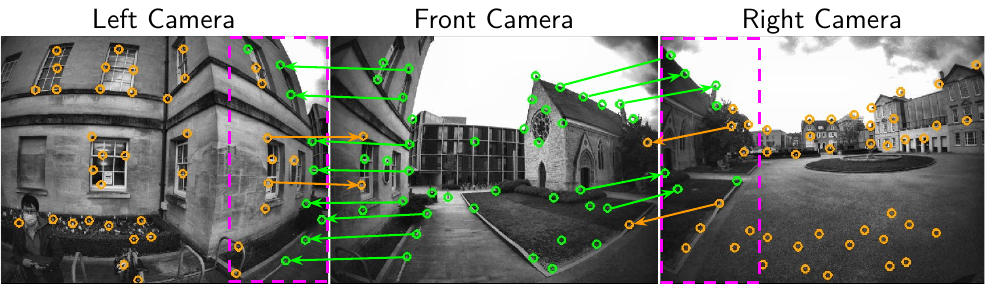}
	\caption{VILENS-MC takes advantage of any overlapping image regions (purple
		rectangles) in a multi-camera setup to track features across cameras.
		This increases feature track length and avoids tracking the same
		feature independently in different cameras. The arrows indicate
		features being tracked from one image to another.}
	\label{fig:cross-cam-tracking}
	\vspace{-1mm}
\end{figure}

\begin{figure}
	\centering
	\includegraphics[width=0.9\linewidth]{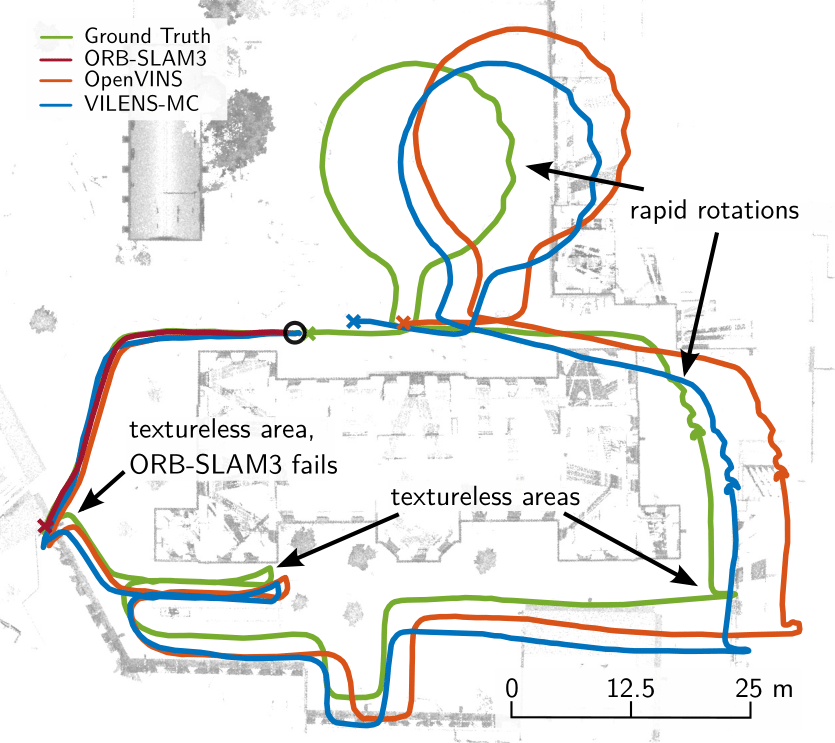}
	\caption{Top-down view of Maths-Hard dataset comparing the estimated
	trajectory of ORB-SLAM3, OpenVINS, and VILENS-MC against the ground truth.
	A black circle marks the start of the trajectory, while colored crosses
	indicate the last pose of each trajectory. A textureless area, where
	multiple cameras were facing towards walls was a cause of failure for the
	stereo odometry methods.}
	\label{fig:math-traj}
\end{figure}

The first is a multi-camera visual-inertial odometry system
(VILENS-MC)~\cite{zhang2021balancing}, developed by the authors. VILENS-MC is
based on factor graph optimization which estimates motion by using all cameras
simultaneously while retaining a fixed overall feature budget. VILENS-MC
introduced cross camera feature tracking as shown in
\figref{fig:cross-cam-tracking}, and focuses on motion tracking in challenging
environments by leveraging this multi-camera dataset. Many sequences in this
dataset, such as Maths-Hard, Quad-Hard or Stairs, would cause classic stereo
inertial odometry approaches to fail. As an example, \figref{fig:math-traj}
shows the trajectories estimated by OpenVINS \cite{Geneva2020},
ORBSLAM-3\cite{Campos2021orbslam3} and VILENS-MC
for the Maths-Hard dataset.

\section{Conclusion}
\label{sec:conclusion}
In this paper, we presented an extension of Newer College Vision and LiDAR
dataset. By leveraging a highly accurate and detailed prior map, we provided
accurate high-frequency 6 DoF ground truth poses, which
distinguishes our dataset from existing ones. We combined a cutting edge
multi-camera device and a dense 3D LiDAR sensor to provide challenging
scenarios, which are considered difficult for current robotic navigation
systems.
The paper also illustrated an example use case of multi-camera visual inertial
odometry. We hope this dataset can propel researchers to further push the
boundaries of autonomous navigation by demonstrating how algorithms can become
more robust when taking into account challenging scenarios.

\section*{Acknowledgment}
The authors would like to thank the members of the Oxford Robotics Institute
(ORI) who helped with the creation of this dataset release, especially Wayne
Tubby. We also thank the personnel of New College for
facilitating our data collection.

This research was supported by the
Innovate UK-funded ORCA Robotics Hub (EP/R026173/1) and the EU H2020 Project THING.
Maurice Fallon is supported by a Royal Society University Research Fellowship.

\bibliographystyle{IEEEtran}
\balance
\bibliography{library}

\end{document}